\documentclass[conference]{IEEEtran}
\usepackage{graphicx} 
\usepackage{amsmath,amsfonts}
\usepackage{graphicx}
\usepackage{textcomp}
\usepackage{xcolor}
\usepackage[utf8]{inputenc}
\usepackage{amsfonts}
\usepackage{enumitem}
\usepackage{caption}
\usepackage{subcaption}
\usepackage[export]{adjustbox}
\usepackage{bbding}
\usepackage{makecell}
\usepackage{float}
\usepackage{amsmath}
\usepackage{algorithm}
\usepackage{url}
\usepackage{hyperref}
\usepackage[noend]{algpseudocode}

\newcommand{\mA}{\mathbf{A}}

\newcommand{\mF}{\mathbf{F}}

\newcommand{\mM}{\mathbf{M}}
\newcommand{\mP}{\mathbf{P}}
\newcommand{\mQ}{\mathbf{Q}}

\newcommand{\mX}{\mathbf{X}}
\newcommand{\mY}{\mathbf{Y}}

\newcommand{\vB}{\mathbf{b}}

\newcommand{\vF}{\mathbf{f}}

\newcommand{\vS}{\mathbf{s}}
\newcommand{\vZ}{\mathbf{z}}

\title{Scaling Graph Neural Networks for Particle Track Reconstruction}

\author{\IEEEauthorblockN{
     Alok Tripathy\IEEEauthorrefmark{1},
     Alina Lazar\IEEEauthorrefmark{3},
     Xiangyang Ju\IEEEauthorrefmark{2},
     Paolo Calafiura\IEEEauthorrefmark{2},
     Katherine Yelick\IEEEauthorrefmark{1},
     Ayd{\i}n Bulu\c{c}\IEEEauthorrefmark{3}}    \vspace{0.5em}
   \IEEEauthorblockA{\IEEEauthorrefmark{1}
     University of California, Berkeley} 
   \IEEEauthorblockA{\IEEEauthorrefmark{2}
     Lawrence Berkeley National Laboratory}
   \IEEEauthorblockA{\IEEEauthorrefmark{3}
     Youngstown State University}
}  
\begin{document}

\maketitle

\begin{abstract}
Particle track reconstruction is an important problem in high-energy physics (HEP), necessary to study properties of subatomic particles. Traditional track reconstruction algorithms scale poorly with the number of particles within the accelerator. The Exa.TrkX project, to alleviate this computational burden, introduces a pipeline that reduces particle track reconstruction to edge classification on a graph, and uses graph neural networks (GNNs) to produce particle tracks. However, this GNN-based approach is memory-prohibitive and skips graphs that would exceed GPU memory. We introduce improvements to the Exa.TrkX pipeline to train on samples of input particle graphs, and show that these improvements generalize to higher precision and recall. In addition, we adapt performance optimizations, introduced for GNN training, to fit our augmented Exa.TrkX pipeline. These optimizations provide a $2\times$ speedup over our baseline implementation in PyTorch Geometric.
\end{abstract}
\section{Introduction}
Charged particle track reconstruction is an important but computationally expensive problem in high-energy physics (HEP). In the ATLAS experiment, proton-proton collisions within the Large Hadron Collider (LHC) accelerator produces a large number of charged particles~\cite{aad2008atlas}. Detectors within the LHC record 3D coordinates of each particle's location when it hits a detector, known as \textit{hit} points, and track reconstruction algorithms connect these 3D coordinates to produce the paths each particle undertook in the accelerator. 

Traditional reconstruction algorithms scale superlienarly with the number of collisions and are specific to the detector hardware~\cite{ATLAS:2017kyn,cms2014description}. By contrast, the Exa.TrkX collaboration developed a pipeline that reduces particle track reconstruction to an edge classification problem, efficiently constructing particle tracks with Graph Neural Networks (GNNs)~\cite{Ju2021-rv}. This GNN-based approach offers a way to construct particle tracks that scales with the number of hit points, and generalizes to different detector hardwares. However, GNNs can be prohibitive on memory-constrained GPU devices when the input graph is large. The Exa.TrkX pipeline circumvents this issue by skipping graphs that would exceed GPU memory when trained. However, this hurts generalization for two reasons: 1) training on full particle graphs is equivalent to training on large batches, which generalizes worse than small batch training~\cite{keskar2017on}, and 2) skipping particle graphs reduces the total amount of training data that the model could learn from. We introduce an approach that trains the particle tracking GNN step of Exa.TrkX with a minibatch approach on vertex subsets of the original particle graphs, rather than whole particle graphs.

However, training on vertex batches in GNNs requires running a computationally expensive sampling step. Training a batch with an $L$-layer GNN requires touching the entire $L$-hop neighborhood of the vertices in the batch. Since the $L$-hop neighborhood is frequently too large to fit in GPU memory, GNN training requires sampling from the $L$-hop neighrborhood. These sampling algorithms, broadly categroized into node-wise, layer-wise, and subgraph sampling algorithms, frequently take up to $60\%$ of the total training time~\cite{jangda2021accelerating}.

In this work, we offer two main contributions. First, we show that a minibatch approach that uses subgraph-sampling generalizes to higher precision and recall than the full-graph approach used in the current Exa.TrkX pipeline. Second, we offer performance optimizations to speedup training. We first adapt an approach to accelerate GNN sampling,  introduced by Tripathy et. al. for node-wise and layer-wise sampling, to subgraph sampling in our augmented pipeline. Our second optimization accelerates the \textit{all-reduce} call used in distributed data parallelism. Altogether, these optimizations speed up training by $2\times$ compared a minibatch implementation with PyTorch Geometric (PyG)~\cite{pyg}.
\section{Background}
\subsection{Particle Track Reconstruction}
\begin{figure}[!t]
    \centering
    \includegraphics[scale=0.25]{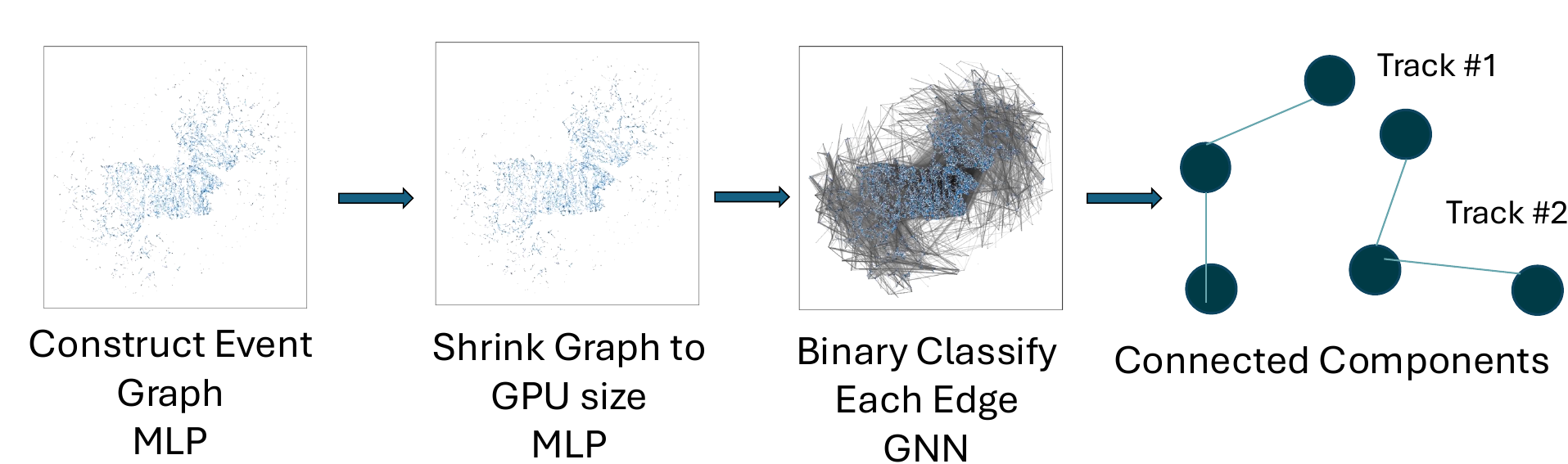}
    \caption{Exa.TrkX GNN pipeline}
    \label{fig:exatrkx-pipeline}
    \vspace{-0.5cm}
\end{figure}

At a high level, the Exa.TrkX pipeline builds a graph per collision event, where each vertex is a 3D coordinate and each edge connects coordinates if they could plausibly be adjacent in a particle's path. Once the pipeline constructs these graphs, a GNN runs binary edge classification to categorize edges as either a member of a true particle track or not a part of a track. The final result, when removing edges from $G$ not in particle tracks, are connected components that represent each particle's track within the accelerator. 

The pipeline consists of five stages. First, it embeds each 3D coordinate with a multi-layer perceptron (MLP). The MLP maps coordinates belonging to the same track near each other in the embedding space. Second, the pipeline builds a fixed-radius nearest neighbor graph from the embedding space. The output \textit{particle graph} has one vertex per coordinate, with edges connecting nearby coordinates within the embedding space. Third, the pipeline shrinks this graph with an MLP before being fed into the memory-intensive GNN. This MLP classifies some edges as \texttt{false} (i.e., not a part of a particle track) and removes them from the graph. The output graph is then fed into a GNN to classify each remaining edge and remove edges not classified as a particle track edge. Lastly, the pipeline runs a connected components algorithm on the graph to construct the candidate particle tracks. Each MLP and GNN in the pipeline must be trained on a set of particle graphs before running inference.
\subsection{Graph Neural Networks}
Graph Neural Networks (GNNs) take as input a graph $G = (V, E)$, node feature data $\mX \in \mathbf{R}^{n\times f}$, and edge feature data $\mY \in \mathbf{R}^{m\times f}$. While GNNs can solve a wide variety of machine learning problems, we focus on \textit{edge classification} in this discussion since particle track reconstruction reduces to edge classification. Each vertex and each edge take an associated \textit{feature vector} as input, and a subset of edges have an associated \textit{label}. The objective of the network is to classify unlabeled edges in the graph using input features, graph connectivity, and labels. 

GNNs follow the \textit{message-passing} model, consisting of a \texttt{message} step and an \texttt{aggregate} step per iteration of training~\cite{hamiltonInductive2017}. The \texttt{message} step creates a message per edge in the graph. The \texttt{aggregate} step takes a vertex $v$ and combines the messages across all of $v$'s incoming neighbors. The output is then passed into an MLP $\phi$ and an activation $\sigma$, and the result is an embedding vector $z_v$. After several layers of these steps, the network outputs an embedding vector per vertex, after which the network inputs vectors and labels into a loss function for backpropagation. Formally, for an arbitrary layer $l$, computing an embedding vector $z^l_v$ is
$$\vZ^l_v = \text{AGG}(\text{MSG}(\vZ^{l-1}_u, \vZ^{l-1}_v)) \forall u\in N(v), 
\vZ^l_v = \sigma(\phi^l(\vZ^l_v))$$
To compute the embedding for an edge $e = (u, v)$, we pass the concatenation of its endpoint vertices feature vectors into an MLP.
$$\vZ^l_e = \phi([\vZ^l_u\text{   }\vZ^l_v])$$

Minibatches in GNN training are subsets of vertices from the input graph $G$. While the size of the batch is small compared to the vertex set, running message-passing on a batch naively touches a large fraction of the input graph. Training a batch $B$ in an $L$-layer network accesses the $L$-hop aggregated neighborhood of $B$. This phenomenon is referred to as \textit{neighborhood explosion}. To alleviate costs, minibatch training GNNs includes a \textit{sampling step} that samples the $L$-hop neighborhood of each batch. The specific sampling algorithm depends on the problem, and there are many such algorithms in the literature, broadly categorized into node-wise, layer-wise, and subgraph sampling algorithms~\cite{hamiltonInductive2017,ladies,graphsaint,shadow}. In addition, in our notation, the $L$-th layer contains the batch vertices, and the 1st layer has vertices furthest from the batch vertices.

\subsection{Distributed Data Parallelism}
We use distributed data parallelism (DDP) to train our GNN across multiple GPUs. In DDP, each batch is split across $P$ GPUs, and GPUs train their shard of the batch in parallel. Once training a batch is complete, we call run an \textit{all-reduce} on each parameter weight matrix to synchronize the model gradients, and continue training the next batch.
\section{Methodology}
\subsection{Interaction GNN}
The Exa.TrkX pipeline uses an Interaction Network (IGNN) as its GNN model~\cite{ignn}. Each layer of an IGNN is one iteration of message passing. The message step for each edge concatenates its edge feature vector with its endpoint vertices' feature vectors, and passes the result through an MLP. Each vertex then sums its incoming messages as part of aggregation, and embeds the resulting sum with an MLP. In addition, each layer must store all the output matrices it constructs ($\mX^{l+1}, \mY^{l+1}, \mM_{src},\mM_{dst}$) in memory, as these matrices will be read in backpropagation. Algorithm~\ref{alg:ignnlayer} details this process in greater detail.

\begin{algorithm}[t]
\begin{algorithmic}[1]
\Procedure{IGNNLayer}{$\mA, \mX, \mY$}
\State $\mX^0 \leftarrow \phi(\mX)$
\State $\mY^0 \leftarrow \phi(\mY)$
\For{$l = 0$ to $L$} \Comment{iterate over each layer}
    \State $\mX' \leftarrow [\mX^l\hspace{0.5em}\mX^0]$
    \State $\mY' \leftarrow [\mY^l\hspace{0.5em}\mY^0]$
    \State $\mY^{l+1} \leftarrow \phi([\mY\hspace{0.5em}\mX[\mA.rows]\hspace{0.5em}\mX[\mA.cols]])$ \Comment{MSG }
    \State $\mM_{src} \leftarrow \Call{REDUCTION}{\mY, \mA.rows, +}$ \Comment{AGG}
    \State $\mM_{dst} \leftarrow \Call{REDUCTION}{\mY, \mA.cols, +}$ \Comment{AGG}
    \State $\mX^{l + 1} \leftarrow \phi([\mM_{src}\hspace{0.5em}\mM_{dst}\hspace{0.5em}\mX])$
\EndFor
\State \Return $\phi(\mY^{L-1})$
\EndProcedure
%
\end{algorithmic}
\caption{Forward pass for Interaction GNN layer. Inputs to the forward pass are the adjacency matrix for the graph $\mA\in \{0,1\}^{n\times n}$ stored in COO format, node features $\mX\in\mathbb{R}^{n\times f}$, and edge features $\mY\in\mathbb{R}^{m\times f}$. While each MLP (denoted by $\phi$) is distinct, superscripts are omitted for simplicity.} \label{alg:ignnlayer}
\vspace{-0.2cm}
\end{algorithm}

\subsection{Minibatch Sampling}
\subsubsection{Matrix-Based ShaDow Implementation}
\begin{figure*}[!t]
    \centering
    \includegraphics[scale=0.40]{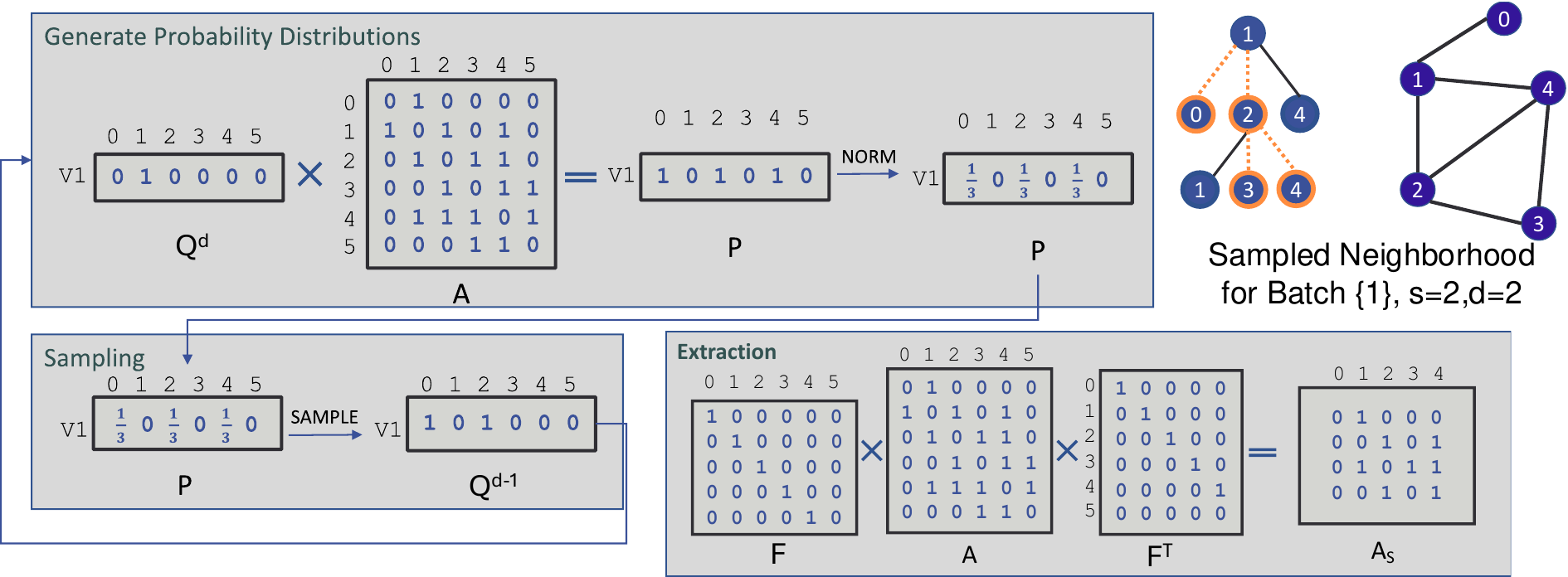}
    \caption{Matrix-Based ShaDow Sampling Algorithm for the example graph and batch. When sampling multiple minibatches, we would stack the $\mQ^d$ matrices for each batch and input the output stacked matrix into the bulk-sampling routine.}
    \label{fig:shadow}
    \vspace{-0.5cm}
\end{figure*}
The current Exa.TrkX pipeline inputs entire particle graphs into the GNN when training, which we refer to as \textit{full-graph} training. That is, the input to an IGNN layer is the entire adjacency matrix for a particle graph. While this approach offers some benefits, such as simplicity and a smoother gradient descent, there are several drawbacks. First, the input graph is oftentimes large. Training large particle graphs will sometimes exceed GPU memory, due to the high overhead of storing intermediate activations per edge in the graph for each layer. Exa.TrkX will skip particle graphs that are too large to be trained. Second, minibatch stochastic gradient descent with small batches will oftentimes converge better than full-batch gradient descent because of additional noise~\cite{keskar2017on}. These limitations motivated us to study minibatch approaches for the Exa.TrkX pipeline.

Minibatches in GNN training are subsets of vertices in the original input graph. Because of the neighborhood explosion problem, where the $L$-hop neighborhood of batch of vertices is prohibitively large, GNN algorithms will run a sampling algorithm on the $L$-hop neighborhood. In this work, we use ShaDow as our sampling algorithm~\cite{shadow}. 

ShaDow samples a subgraph $\mA_S$ for each vertex in a minibatch, and inputs $\mA_S$ into forward propagation as opposed to $\mA$. The algorithm runs a separate random walk for each batch vertex, sampling with fanout $s$ and depth $d$ as defined by the user, and returns the induced subgraph of all vertices touched in the random walk. A batch with $b$ vertices will output a single sampled adjacency matrix $\mA_S$ with $b$ distinct components, one for each batch vertex's induced subgraph. The first procedure in Algorithm~\ref{alg:shadow} details this process.

\begin{algorithm}[t]
\begin{algorithmic}[1]
\Procedure{ShaDow}{$\mA, \vB$}
\State $\mA_S \leftarrow \emptyset$
\For{$b \in \vB$}
    \State $\vS \leftarrow []$ \Comment{vertices touched during random walk}
    \State $\vF \leftarrow [b]$
    \For{$i = 0\ldots d$} \Comment{$d$ is depth of random walk}
        \State $\vF' \leftarrow$ $s$ distinct neighbors for each vertex in $\vF$
        \State $\vS \leftarrow \vS + \vF'$
        \State $\vF \leftarrow \vF'$
    \EndFor
   \State $\mA_S' \leftarrow$ \Call{SUBGRAPH}{$\mA, \vF$}
   \State $\mA_S \leftarrow$ \Call{APPEND\_COMPONENT}{$\mA_S, \mA_S'$}
\EndFor
\State \Return $\mA_S$
\EndProcedure
\vspace{-0.02cm}

\end{algorithmic}
\caption{Standard implementation of ShaDow sampling algorithm. This algorithm accepts as input the adjacency matrix $\mA$ and batch vertices a vector $\vB$.} \label{alg:shadow}
\vspace{-0.1cm}
\end{algorithm}
\subsection{Performance Optimizations}
In addition to adding minibatch training, we introduce performance optimizations to our augmented Exa.TrkX pipeline. First, we accelerate our implementation for ShaDow sampling with \textit{matrix-based bulk sampling} as introduced in Tripathy et. al.~\cite{tripathy2024distributed}. GNN sampling takes roughly 50\% of the total GNN training time in the Exa.TrkX pipeline, and matrix-based bulk sampling accelerates GNN sampling by representing sampling with matrix operations. This approach allows the pipeline to sample multiple minibatches in a single training step rather than sampling each batch successively, increasing GPU utilization. However, while matrix-based bulk sampling theoretically supports any sampling algorithm, it was only concretely introduced with node-wise and layer-wise sampling algorithms~\cite{hamiltonInductive2017,ladies}. We show how to accelerate ShaDow, a subgraph sampling algorithm, within the matrix-based bulk sampling framework.

There are three main steps to matrix-based sampling, which are outlined in Figure~\ref{fig:shadow} in the case of ShaDow. First, we run $\mQ^{d-1} \leftarrow \mQ^d\mA$, with a matrix $\mQ^d$ that we construct. For one ShaDow batch, $\mQ^d$ has size $b\times n$ and has one non-zero per row at the column index of each batch vertex. Multiplying $\mQ^d\mA$ constructs a matrix with the neighborhood of each batch vertex in each row, and dividing each row by its sum results in a uniform probability distribution to sample from. Once we sample $s$ neighbors per batch vertex, we expand $\mQ^{d-1}$ to have one nonzero per row and repeat the sampling process. In addition, we keep track of all vertices touched during sampling in a $b\times n$ sparse matrix $\mF$ for extraction. Once we run $d$ iterations of sampling, matrix-based sampling will run row and column selection SpGEMMs to extract an induced subgraph of $\mA$ per batch vertex, according to the nonzeros in that vertex's row in $\mF$. The output is a sampled adjacency matrix $\mA_S$ with one component per batch vertex.

To sample multiple minibatches with matrix-based sampling, we simply stack the individual $\mQ$ matrices per batch. Across $k$ batches, the $\mQ$, $\mF$, $\mP$ matrices become
\begin{equation}
\mQ^l = \left( 
\begin{array}{c}
\mQ^l_{1} \\
\vdots \\
\mQ^l_{k}
\end{array} 
\right)
\mF = \left( 
\begin{array}{c}
\mF_{1} \\
\vdots \\
\mF_{k}
\end{array} 
\right)
\mP = \left( 
\begin{array}{c}
\mP_{1} \\
\vdots \\
\mP_{k}
\end{array} 
\right)
\end{equation}
\subsection{Distributed All-Reduce}
In addition to using matrix-based sampling, we optimize the \textit{all-reduce} call when training with DDP. An IGNN holds multiple separate parameter matrices because of the separate MLPs used in training, most of which have dimension $f\times f$. Running separate \textit{all-reduce} reductions on each parameter matrix yields high latency costs. We instead stack these parameter matrices and run a single \textit{all-reduce} call, which significantly lowers the latency cost of the \textit{all-reduce}.
\section{Results}
\subsection{Experimental Setup}
We run all our experiments on the Perlmutter supercomputer at NERSC. Perlmutter GPU nodes have a  AMD EPYC 7763 (Milan) CPU and four NVIDIA A100 GPUs. Each pair of the GPUs has NVLINK 3.0 to communicate data at $100$GB/s unidirectional bandwidth. We implement our framework in PyTorch 2.3.1~\cite{paszke2019pytorch} and PyTorch Geometric (PyG) 2.5.3 with CUDA 12.2~\cite{pyg}. We use NCCL 2.21 as our communications library~\cite{ncclRepo}. Our baseline is the existing Exa.TrkX pipeline in the acorn repository with PyG's implementation of the ShaDow algorithm. 

We run experiments on two HEP datasets: Connect the Dots (CTD) and Example 3 (Ex3), which are summarized in Table~\ref{tab:datasets}~\cite{acornrepo}. Each dataset is a set of disjoint event graphs, and we use 80 graphs for training, 10 for validation, and 10 for testing. While CTD offers $7800$ total training graphs, we use $80$ as a representative sample to show the effectiveness of our tool. We train all our datasets with a batch size of $256$ a hidden dimension of $64$, and $30$ epochs. Our hyperparameters for ShaDow are depth $d=3$ and fanout $s=6$, and our GNN is trained on $8$ layers (i.e., $8$ iterations of message-passing).
\begin{table}[t]
\centering
\caption{Datasets used in our experiments}
\label{tab:data}
\begin{tabular}{ |l|r|r|r|r|r|r|}
\hline
Name & Graphs & \makecell{Avg \\ Vertices} & \makecell{Avg \\ Edges} & \makecell{MLP \\ Layers} & \makecell{Vertex \\ Features} & \makecell{Edge \\ Features} \\
\hline
CTD & 80 & 330.7K & 6.9M & 3 & 14 & 8 \\
Ex3 & 80 & 13.0K & 47.8K & 2 & 6 & 2 \\
\hline
\end{tabular}
\label{tab:datasets}
\vspace{-0.8cm}
\end{table}
\subsection{Convergence Results}
\begin{figure}[t]
    \centering
    \includegraphics[scale=0.05]{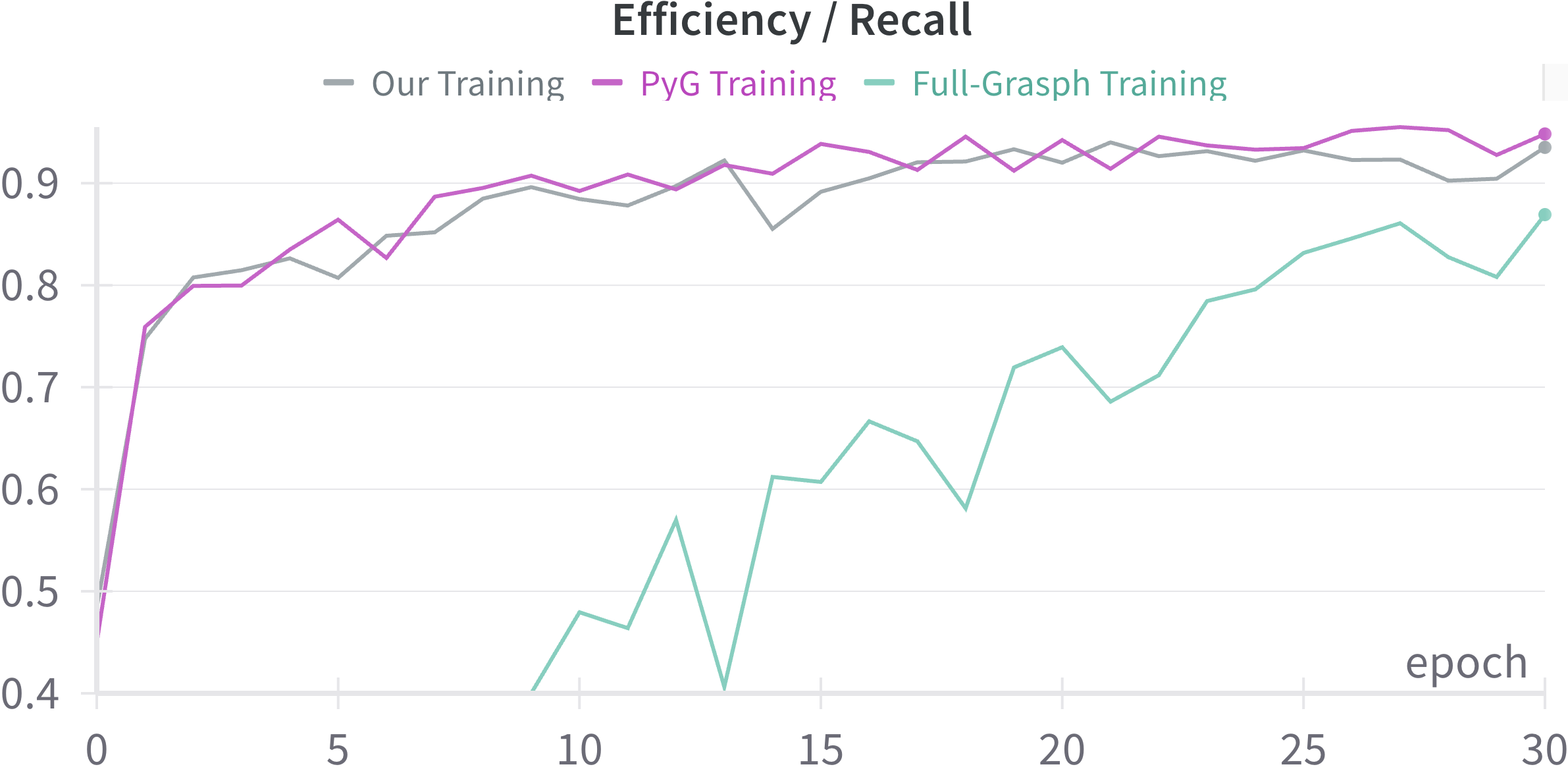}
    \includegraphics[scale=0.05]{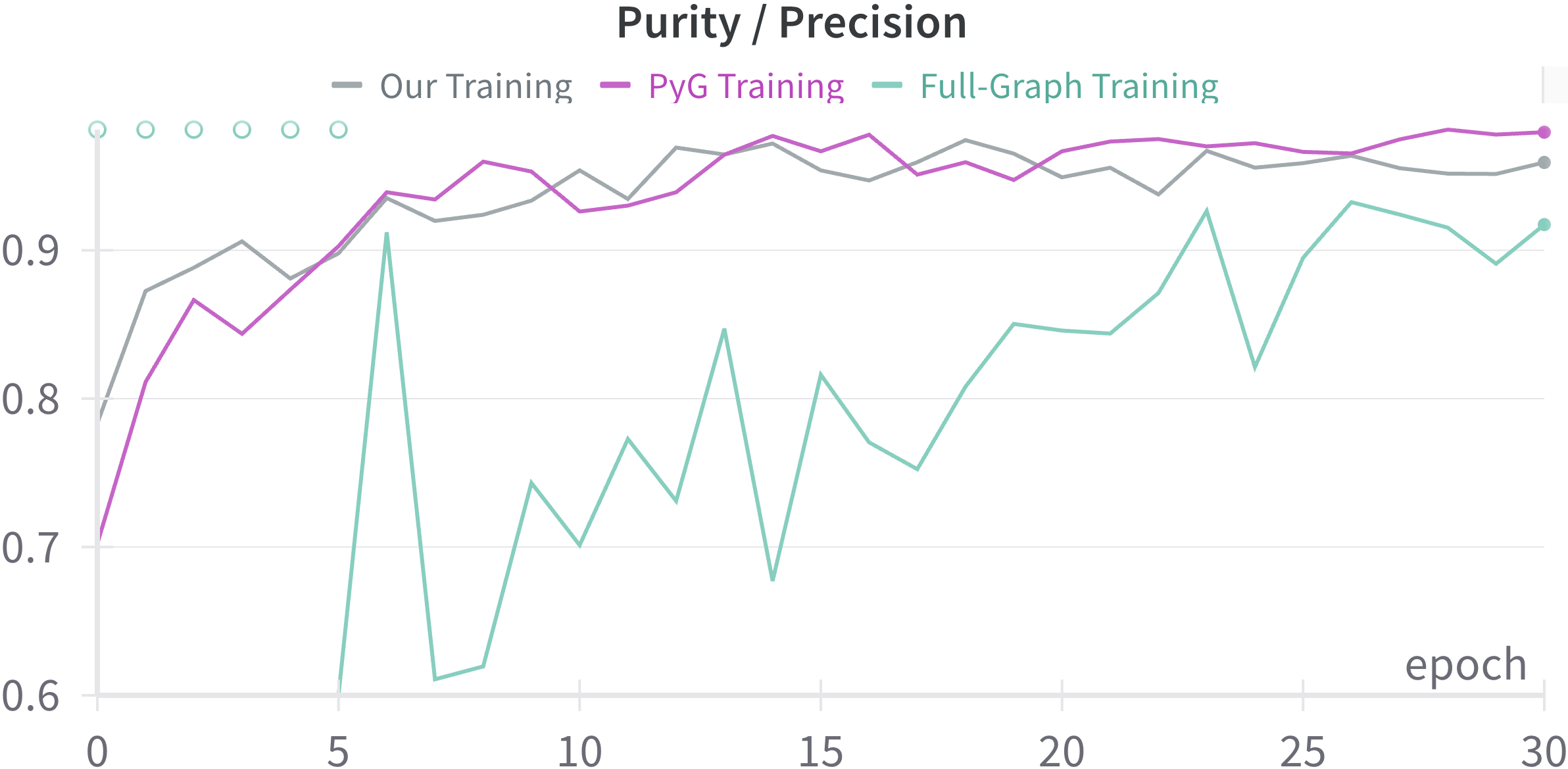}
    \caption{Convergence results on Ex3 for full-graph training, ShaDow training with PyG's implementation, and ShaDow training with our implementation. Precision and recall are based on the number of correctly classified edges across validation set particle graphs and the total number of edges.}
    \label{fig:convergence-results}
    \vspace{-0.7cm}
\end{figure}

We compare the precision and recall convergence of our minibatch-based GNN training pipeline with the original Exa.TrkX pipeline in Figure~\ref{fig:convergence-results}. The original pipeline feeds entire particle graphs as input into the GNN, with $80$ training steps per epoch with our 80/10/10 training split. This can be considered an instance of large-batch training, where the batch size is the average size of a particle graph. While training on entire particle graphs is simpler and avoids running a sampling algorithm, it comes with two main drawbacks. First, large-batch training frequently converges to lower accuracy than smaller minibatch training~\cite{keskar2017on}. Sampling subgraphs with a tunable batch size allows for training on smaller batch sizes and potentially better generalization. Second, the original pipeline must skip graphs that are too large to train in GPU memory. This is because training requires storing large activation matrices per layer of the GNN, the largest of which have $mf$ total elements. When $m$ for a graph is large, training that graph can exceed GPU memory. Figure~\ref{fig:convergence-results} shows that our minibatch approach with batch size $256$ converges to a higher precision and recall than the full-graph approach. In addition, the PyG convergence curves compared to our implementation's convergence curves show that our approach does not suffer from precision or recall degradation. 
\subsection{Epoch Time Results}
\begin{figure}[t]
    \centering
    \includegraphics[scale=0.23]{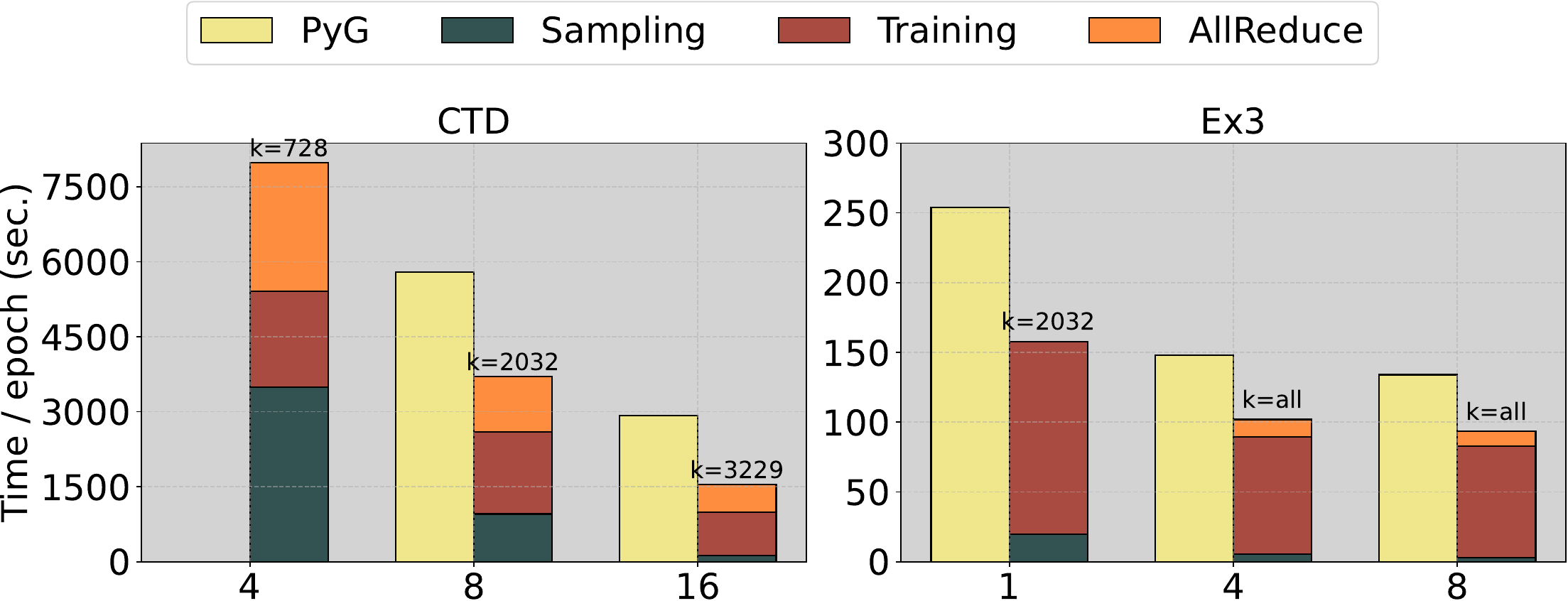}
    \caption{Epoch Time results for the Exa.TrkX pipeline across GPUs with the PyG implementation of ShaDow, and with our implementation of ShaDow and all-reduce optimization. Here, $k$ is the number of minibatches that were sampled in bulk during a single step of training. PyG timed out on $p=4$ processes for CTD training, so results could not be collected.}
    \label{fig:epoch-time-results}
    \vspace{-0.5cm}
\end{figure}

We compare the time to train one epoch between our pipeline with PyG's ShaDow implementation and our ShaDow implementation in Figure~\ref{fig:epoch-time-results} across process counts. We distribute training with distributed data parallelism, where each process uses a local batch size of $256/P$ on $P$ processes. We see a roughly $1.3\times$ to $2\times$ speedup compared to PyG's implementation across both datasets. This is largely attributed to increased GPU utilization with our sampling approach, and decreased latency costs from our coalesced \textit{all-reduce} call. On CTD, note that the sampling costs scale superlinearly, roughly $27\times$ with $4\times$ more GPUs. This is because, in addition to sampling time scaling linearly with DDP, our approach is able to sample more minibatches in bulk as we increase the number of GPUs due to increased aggregate memory. In addition, the training time scales $2.2\times$ faster with $4\times$ more GPUs. For Ex3, the sampling times scale roughly $6.7\times$ faster with $8\times$ more GPUs. Speedups are generally smaller for Ex3, as it is a smaller dataset with less work during training.

\section{Conclusion}
In conclusion, we accelerate GNN training in the original Exa.TrkX particle tracking pipeline. All of our code is available at \href{https://github.com/PASSIONLab/CAGNET}{\color{blue}{https://github.com/PASSIONLab/CAGNET}}.
\section{Acknowledgements}
We thank our reviewers for their constructive insight and feedback. This work is supported by the Advanced Scientific Computing Research (ASCR) Program of the Department of Energy Office of Science under contract No. DE-AC02-05CH11231. This research used resources of the National Energy Research Scientific Computing Center (NERSC), a Department of Energy Office of Science User Facility using NERSC award ASCR-ERCAP0033069.

\bibliography{cagnet-exatrkx}
\bibliographystyle{plain}

\end{document}